\documentclass[11pt,letterpaper]{article}
\usepackage{naaclhlt2013}
\usepackage{times}
\usepackage{latexsym}
\usepackage{float}
\usepackage{graphicx}
\usepackage{bibentry}
\usepackage{amssymb, amsmath}
\usepackage{breqn}
\title{ClaC: Semantic Relatedness of Words and Phrases}

\author{Reda Siblini\\
        Concordia University\\
        1400 de Maisonneuve Blvd. West\\
        Montreal, Quebec, Canada, H3G 1M8\\
	    {\tt r\_sibl@encs.concordia.ca}
	  \And
	    Leila Kosseim\\
  	    Concordia University\\
        1400 de Maisonneuve Blvd. West\\
        Montreal, Quebec, Canada, H3G 1M8\\
        {\tt kosseim@encs.concordia.ca}}

\date{}

\begin{document}
\maketitle
\begin{abstract}
The measurement of phrasal semantic relatedness is an important metric for many natural language processing applications. In this paper, we present three approaches for measuring phrasal semantics,  one based on a semantic network model, another on a distributional similarity model, and a hybrid between the two. Our hybrid approach achieved an F-measure of 77.4\% on the task of evaluating the semantic similarity of words and compositional phrases.
\end{abstract}

\section{Introduction}
Phrasal semantic relatedness is a measurement of how multiword expressions are related in meaning. Many natural language processing applications such as textual entailment, question answering, or information retrieval require a robust measurement of phrasal semantic relatedness. Current approaches to address this problem can be categorized into three main categories: those that rely on a knowledge base and its structure, those that use the distributional hypothesis on a large corpus, and hybrid approaches. In this paper, we propose supervised approaches for comparing phrasal semantics that are based on a semantic network model, a distributional similarity model, and a hybrid between the two. Those approaches have been evaluated on the task of semantic similarity of words and compositional phrases and on the task of evaluating the compositionality of phrases in context.\\

\begin{figure*}
\begin{center}
\caption{Example of the semantic network around the word {\em car}.\label{fig:conceptrelationsexample}}
\includegraphics  [width=0.6 \textwidth]{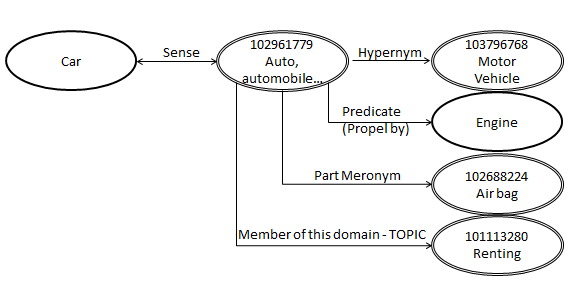}
\end{center}
\end{figure*}

\section{Semantic Similarity of Words and Compositional Phrases}
The semantic similarity of words and compositional phrases is the task of evaluating the similarity of a word and a short phrase of two or more words; for example, the word {\em Interview} and the phrase {\em Formal Meeting}. In the next section we present our semantic network model for computing phrasal semantic relatedness between a word and a phrase, followed by a distributional similarity model, that we evaluate on the task of semantic similarity of words and compositional phrases.

\subsection{Semantic Network Model}

Knowledge-based approaches to semantic relatedness use the features of the knowledge base to measure the relatedness. One of most frequently used semantic network is the Princeton's WordNet~\cite{Fellbaum1998} which groups words into synonyms sets (called synsets) and includes 26 semantic relations between those synsets, including: hypernymy, hyponymy, meronymy, entailment~\ldots\\
To measure relatedness, most of those approaches rely on the structure of the semantic network, such as the semantic link path, depth \cite{Leacock1998a,Wu1994}, direction \cite{Hirst1998}, or type \cite{Tsatsaronis2010}. Our phrasal semantic relatedness approach is inspired from those methods. However, our approach is based on the idea that the combination of the least costly types of relations that relate one concept to a set of concepts are a suitable indicator of their semantic relatedness. The type of relations considered includes not only the hyponym/hypernym relations but also all 26 available semantic relations found in WordNet in addition to relations extracted from each of the eXtended WordNet \cite{harabagiu1999wordnet} synset's logical form. 

\begin{table*}
\begin{center}
\begin{tabular}{|l|r|p{4 in}|}
\hline \bf Category & \bf Weight &\bf Semantic Relations in WordNet or xWordnet\\ 
\hline
$Similar$& $1$ & similar to, pertainym, participle of verb, entailment, cause, antonym, verb group
\\$Hypernym$&$2$ & hypernym, instance hypernym, derivationally related
\\$Sense$& $4$ & lemma-synset
\\$Predicate$& $6$ & predicate (extracted from Extended WordNet)
\\$Part$& $8$ & holonym (instance, member, substance), meronym (instance, member, substance), inverse predicate (extracted from Extended WordNet)
\\$Instance$& $10$ & hyponym, instance hyponym
\\$Other$& $12$ & attribute, also see, domain of synset (topic, region, usage), member of this domain (topic, region, usage)\\ \hline
\end{tabular}
\end{center}
\caption{\label{tb:categories}Relations Categories and Corresponding Weights.}
\end{table*}

To implement our idea, we created a weighted and directed semantic network based on the relations of WordNet and eXtended WordNet. We used WordNet's words and synsets as the nodes of the network. Each word is connected by an edge to its synsets, and each synset is in turn connected to other synsets based on the semantic relations included in WordNet. In addition each synset is connected by a labeled edge to the predicate arguments that are extracted from the eXtended WordNet synset's logical form. Every synset in the eXtended WordNet is related to a logical form, which contains a set of predicate relations that relates the synset to set of words. Each predicate in this representation is added as an edge to the graph connecting the synset to a word. For example, Figure~\ref{fig:conceptrelationsexample} shows part of the semantic network created around the word {\em car}. In this graph, single-line ovals represent words, while double-line ovals represent synsets. 

To compute the semantic relatedness between nodes in the semantic network, it is necessary to take into consideration the semantic relation involved between two nodes. Indeed, WordNet's 26 semantic relations are not equally distributed nor do they contribute equally to the semantic relatedness between concept. In order to indicate the contribution of each relation, we have classified them into seven categories: {\em Similar}, {\em Hypernym}, {\em Sense}, {\em Predicate}, {\em Part}, {\em Instance}, and {\em Other}. By classifying WordNet's relations into these classes, we are able to weight the contribution of a relation based on the class it belongs to, as opposed to assigning a contributory weight to each relations.
The weights were assigned by manually comparing the semantic features of a set of concepts that are related by a specific semantic relations. Table~\ref{tb:categories} shows the seven semantic categories that we defined, their corresponding weight, and the relations they include.  For example the category {\em Similar} includes WordNet's relations of {\em entailment}, {\em cause}, {\em verb group}, {\em similar to}, {\em participle of verb}, {\em antonym}, and {\em pertainym}. This class of relations has the most common semantic features when comparing two concepts related with any of those relations and hence was assigned the lowest weight\footnote{The weight can be seen as the cost of traversing an edge; hence a lower weight is assigned to a highly contributory relation.} of 1. All the 26 relations in the table are the ones found in WordNet, for the exception of the predicate (and inverse predicate) relations which are the predicate relations extracted from the eXtended WordNet. This can be seen in~Figure~\ref{fig:conceptrelationsexample}, for example, where the word {\em car} is related to the word {\em Engine} with the {\em Predicate} relation extracted from the eXtended WordNet logical form and more specifically the predicate {\em propel by}.

The computation of semantic relatedness between a word and a compositional phrase is then  the combination of weights of the shortest weighted path\footnote{The shortest path is based on an implementation of Dijkstra’s graph search algorithm \cite{dijkstra1959note}} in the weighted semantic network between that word and every word in that phrase, normalized by the maximum path cost.\\

\begin{figure*}
\begin{center}
\caption{Shortest Path Between the Word {\em Interview} and the Phrase {\em Formal Meeting}.\label{fig:semanticpathexample}}
\includegraphics  [width=0.6 \textwidth]{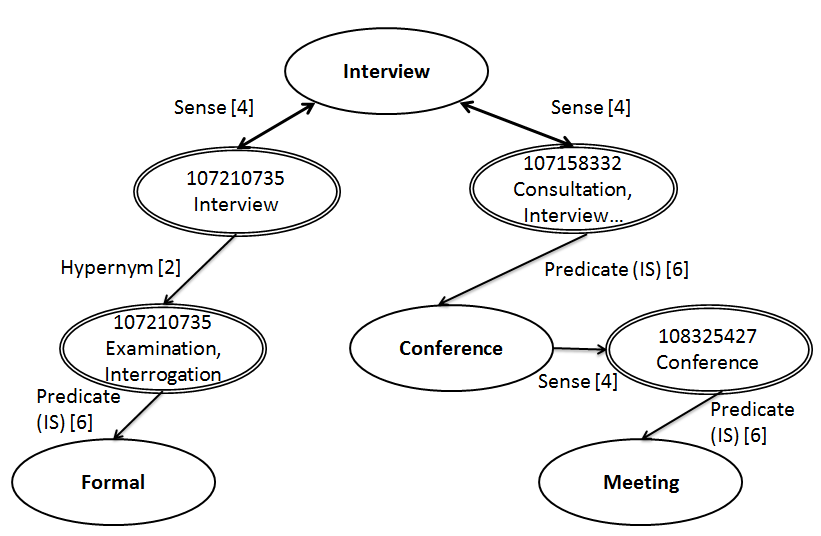}
\end{center}
\end{figure*}

Figure~\ref{fig:semanticpathexample} shows an extract of the network involving the words {\em Interview} and the phrase {\em Formal Meeting}. For the shortest path from {\em Interview} to {\em Formal}, the word {\em Interview} is connected with a {\em Sense} relation to the synset {\em \#107210735 [Interview]}. As indicated in Table 1, the weight of this relation is defined as 4, This synset is connected to the synset {\em Examination} through a {\em Hypernym} relation type with a weight of 2, which is connected to the word {\em Formal} with a predicate (IS) relation of weight 6. 
Overall, the sum of the shortest path from {\em Interview} to {\em Formal Meeting} is hence equal to the sum of the edges shown in Figure~1 (4+2+6+4+6+4+6 = 32). By normalizing the sum to the maximum, In our approach, 24 is maximum path cost after which we assume that two words are not related (which we assume to be traversing two times maximum weighted path, 2 * maximum path weight of 12) and 8 is the minimum number of edges between 2 words (which is equal to traversing from the word to itself, 2 * sense weight of 4)). Taking into consideration the number of words in the phrase, the semantic relatedness will be (24*2 - (32-8*2))/24*2 = 66.7\%. In the next section, we will introduce our distributional similarity model.

\subsection{Distributional Similarity Model}
Distributional similarity models rely on the distributional hypothesis \cite{harris1954distributional} to represent a word by its context in order to compare word semantics. There are various approach for the selection, representation, and comparison of contextual data. Most use the vector space model to represent the context as dimensions in a vector space, where the feature are frequency of co-occurrence of the context words, and the comparison is usually the cosine similarity. To go beyond lexical semantics and to represent phrases, a compositional model is created, some use the addition or multiplication of vectors such as \newcite{mitchell2008vector}, or the use of tensor product to account for word order as in the work of \newcite{widdows2008semantic}, or a more complex model as the work of \newcite{grefenstette2011experimental}. In our model, we are inspired by those various work, and more specifically by the work of \newcite{mitchell2008vector}. The compositional model is based on phrase words vectors addition, where each vector is composed of the collocation pointwise mutual information of the word up to a window of 3 words left and right of the main word. The corpus used to collect the features and their frequencies is the Web 1TB corpus \cite{brants2006web}.
For the {\em Interview} to {\em Formal Meeting} example, the vector of the word {\em interview} is first created from the corpus of the top 1000 words collocating {\em interview} between the window of 1 to 3 words with their frequencies. A similar vector is created for the word {\em Formal} and the word {\em Meeting}, the vector representing {\em Formal Meeting} is then the addition of vector {\em Formal} to vector {\em Meeting}.
The comparison of vector {\em Interview} to vector {\em Formal Meeting} is then the cosine of both vectors.

\subsection{Evaluation}
We evaluated our approaches for word-phrase semantic relatedness on the SemEval task of evaluating phrasal semantics, and more specifically on the sub-task of evaluating the semantic similarity between words and phrases. The task provided an English dataset of 15,628 word-phrases, 60\% annotated for training and 40\% for testing, with the goal of classifying each word-phrase as either positive or negative.
To transform the semantic relatedness measure to a semantic similarity classification one, we first calculated the semantic relatedness of each word-phrase in the training set, and used JRip, WEKA's~\cite{witten1999weka} implementation of Cohen's RIPPER rule learning algorithm ~\cite{cohen1999simple}, in order to learn a set of rules that can differentiate between a positive semantic similarity and a negative one. The classifier resulted in rules for the semantic network model based relatedness that could be summarized as follows: {\em If the semantic relatedness of the word-phrase is over 61\% then the similarity is positive, otherwise it is negative}. 
So for the example {\em Interview} - {\em Formal meeting}, which resulted in a semantic relatedness of 66.7\% in the semantic network approach, it will be classified positively by the generated rule. This method was our first submitted test run to this task, which resulted in a recall of 63.79\%, a precision of 91.01\%, and an F-measure of 75.00\% on the testing set. 

For the second run, we trained the distributional similarity model using the same classifier. This resulted with the following rule that could be summarized as follows: {\em If the semantic relatedness of the word-phrase is over 40\% then the similarity is positive, otherwise it is negative}. It was obvious from the training set that the semantic network model was more accurate than the distributional similarity model, but the distributional model had more coverage. So for our second submitted test run, we used the semantic network approach as the main result, but used the distributional model as a backup approach if one of the words in the phrase was not available in WordNet, thus combining the precision and coverage of both approaches. This method resulted in a recall of 69.48\%, a precision of 86.70\%, and an F-measure of 77.14\% on the testing set. 

For the last run, we used the same classifier but this time we training it using two features: the semantic network model relatedness measure (SN), and the distributional similarity model (DS). This training resulted in a set of rules that could be summarized as follows: {\em  if SN $>$ 61\% then the similarity is positive, else if DS $>$ 40\% then the similarity is also positive, and lastly if SN $>$ 53\% and DS $>$ 31\% then also in this case the similarity is positive, otherwise the similarity is negative}. This was our third submitted test run, which resulted a recall of 70.66\%, a precision of 85.55\%, and an F-measure of 77.39\% on the testing set. \\

\section{Semantic Compositionality in Context}
The semantic compositional in context is the task of evaluating if a phrase is used literally or figuratively in context. For example, the phrase {\em big picture} is used literally in the sentence {\em Click here for a bigger picture} and figuratively in {\em To solve this problem, you have to look at the bigger picture}. \\
Our approach for this task is a supervised approached based on two main components:
first, the availability of the phrases most frequent collocating expressions in a large corpus, and more specifically the top 1000 phrases by frequency in Web 1TB corpus \cite{brants2006web}. For example, for the phrase {\em big picture}, we collect the top 1000 phrases that come before and after the phrase in a corpus, those includes {\em look at the, see the, understand the ....}. If the context contain any of those phrase, then this component returns 1, indicating that the phrase is most probably used figuratively. The intuition is that, the use of phrases figuratively is more frequent than their use in a literal meaning, and hence the most frequent use will be collocated with phrases that indicate this use.\\
The second component, is the phrase compositionality. We calculate the semantic relatedness using the semantic network model relatedness measure, that was explained in Section 2.1, between the phrase and the first content word before it and after it. The intuition here is that the semantic relatedness of the figurative use of the phrase to its context should be different than the relatedness to its literal use. So for the example, the phrase {\em old school} in the context {\em he is one of the old school} versus {\em the hall of the old school}, we can notice that {\em hall} will be more related to {\em old school} than the word {\em one}. This component will result in two features: the relatedness to the word before the phrase (SRB) and the relatedness to word after the phrase in context (SRA).

To combine both componenets, we evaluated our approaches on the data set presented by the SemEval task of evaluating phrasal semantics, and more specifically on the sub task of evaluating semantic compositionality in context. The data set contains a total of 1114 training instances, and 518 test instances. We use the training data and computed the three features (Frequent Collocation (FC), Semantic Relatedness word Before (SRB), and Semantic Relatedness word After (SRA), and used JRip, WEKA's~\cite{witten1999weka} implementation of Cohen's RIPPER rule learning algorithm ~\cite{cohen1999simple} to learn a set of rule that differentiate between a figurative and literal phrase use. This method resulted in a set of rules that can be summarized as follows: {\em if FC is equal to 0 and SRB $<$ 75\% then it is used literally in this context, else if FC is equal to 0 and SRA $<$ 75\%  then it is is also used literally, otherwise it is used figuratively}. This method resulted in an accuracy of 55.01\% on the testing set. 

\section{Conclusion}
In this paper we have presented state of the art word-phrase semantic relatedness approaches that are based on a semantic network model, a distributional model, and a combination of the two. The novelty of the semantic network model approach is the use of the sum of the shortest path between a word and a phrase from a weighted semantic network to calculate word-phrase semantic relatedness. We evaluated the approach on the SemEval task of evaluating phrasal semantics, once in a supervised standalone configuration, another with a backup distributional similarity model, and last in a hybrid configuration with the distributional model.
The hybrid model achieved the highest f-measure in those three configuration of 77.4\% on the task of evaluating the semantic similarity of words and compositional phrases.
We also evaluated this approach on the subtask of evaluating the semantic compositionality in context with less success, and an accuracy of of 55.01\%.

\section*{Acknowledgments}

We would like to thank the reviewers for their suggestions and valuable comments.

\bibliographystyle{naaclhlt2013}

\end{document}